\documentclass[lettersize,conference]{ieeeconf}

\IEEEoverridecommandlockouts                              
                                                          
\usepackage{cite}
\usepackage{amsmath,amssymb,amsfonts}
\usepackage{algorithmic}
\usepackage{graphicx}
\usepackage{textcomp}
\usepackage{xcolor}
\def\BibTeX{{\rm B\kern-.05em{\sc i\kern-.025em b}\kern-.08em
    T\kern-.1667em\lower.7ex\hbox{E}\kern-.125emX}}
    
\usepackage{xspace}
\let\OldTexttrademark\texttrademark
\renewcommand{\texttrademark}{\OldTexttrademark\xspace }%

\usepackage [english]{babel}
\usepackage [autostyle, english = american]{csquotes}
\MakeOuterQuote{"}
    
\usepackage{hyperref}
\newcommand{\email}[1]{\href{mailto:#1}{#1}}

\usepackage{bm}

\usepackage{cleveref}
\usepackage{afterpage}
\usepackage{gensymb}
\usepackage{paralist}
\usepackage{multirow}
\usepackage{booktabs}
\usepackage{graphicx}
\usepackage{subcaption}  
\usepackage{float}
\usepackage{censor}

\DeclareMathOperator*{\argmin}{argmin}

\begin{document}


\title{Control Lyapunov Functions for Underactuated Soft Robots}

\author{Huy Pham$^{1}$, Zach J. Patterson$^{1}$
\thanks{$^{1}$ Mechanical and Aerospace Engineering, Case Western Reserve University. \email{zpatt@case.edu}}%
\thanks{*This work has been submitted to the IEEE for possible publication. Copyright may be transferred without notice, after which this version may no longer be accessible}%
}

\maketitle

\begin{abstract}
Soft and soft–rigid hybrid robots are inherently underactuated and operate under tight actuator limits, making task-space control with stability guarantees challenging. Common nonlinear strategies for soft robots (e.g., those based on PD control) often rely on the assumption of full actuation with no actuator limits. This paper presents a general control framework for task-space regulation and tracking of underactuated soft robots under bounded inputs. The method enforces a rapidly exponentially stabilizing control Lyapunov function as a convex inequality constraint while simultaneously satisfying underactuated full-body dynamics and actuator bounds. We validate the approach in simulation on several platforms spanning increasing underactuation: a simple two link tendon-driven ``finger'', a trimmed helicoid manipulator, and a highly underactuated spiral robot. We compare against a number of baseline methods from the literature. Results show improved task-space accuracy and consistent Lyapunov convergence under input limits, achieving superior set-point and trajectory-tracking performance.
\end{abstract}

\section{Introduction}
Soft robots and soft-rigid hybrid robots are a growing paradigm due to their inherent compliance and the resulting potential for safety and adaptivity in interactions with unstructured environments. Unlike traditional rigid-link manipulators, soft robots exhibit distributed elasticity and strong coupling between geometry, material properties, and actuation \cite{satina2023model, chen2025survey}. These characteristics enable rich and adaptive behaviors, but they also introduce substantial challenges in modeling and control \cite{satina2023model}.

Controlling soft robots remains difficult due to their inherently underactuated dynamics. Early efforts have largely relied on fully actuated approximations, where the number of control inputs is assumed to match the number of configuration variables \cite{dixon_underactuated_2003}. Under this assumption, classic nonlinear control methods can be extended to soft systems. For example, proportional–derivative (PD)-like control strategies have been used in \cite{cao_model-based_2021, wu_fem-based_2021, george_thuruthel_first-order_2020}, and feedback linearization has been implemented in \cite{kapadia_empirical_2014, falkenhahn_model-based_2015, katzschmann_dynamic_2019, della_santina_model-based_2020, doroudchi_configuration_2021, fischer_dynamic_2022}. 

Fully actuated approximations, however, constitute a clear oversimplification of the true control problem \cite{satina2023model}. Most soft robots are inherently underactuated, as their high number of (or infinite) degrees of freedom cannot be matched by a comparable number of actuators. As a result, not every configuration is achievable as an equilibrium, and control authority cannot be independently exerted along all deformation directions. Previous approaches extend upon the classic model-based joint-space \cite{pustina2022feedback,patterson2024modeling, borja2022energy} or task-space PD+ control approaches \cite{santina2019exact}, while usually ignoring actuator saturation and other constraints. Input constraints fundamentally alter the closed-loop stability properties of dynamical systems and can invalidate the convergence guarantees of classic control laws \cite{hu2001control}. The solution to this issue draws inspiration from optimization-based control, where an objective function governs the desired behaviors of the robot is designed subject to a set of constraints that respect physically feasible conditions. Previous works have dealt with input saturations, but typically not in the underactuated context \cite{santina2019dynamic,shao2023model}.

Optimization-based control is a powerful and flexible framework for handling actuator limits and ensuring stability by computing a sequence of control inputs that minimizes the sum of a control Lyapunov function evaluated over future steps \cite{powell2015model}. MPC has been applied to soft manipulators for trajectory tracking \cite{spinelli2022unified, zou_mpc-based_2025}. Although MPC accommodates underactuated dynamics and actuator constraints, its computational burden is substantial compared to the other methods, especially for soft robots which require a large number of  DOFs to approximate continuum structures. Additionally, getting MPC to run in real-time on nonlinear problems often requires robot-specific simplifications and/or linearization, sacrificing some of the predictive capability and generality of the approach.

A promising framework that balances computational complexity and performance for underactuated systems is the Control Lyapunov Function (CLF)-based Quadratic Program (QP). This framework forces exponential convergence in the sense of Lyapunov together with actuator constraints within a QP in each control step \cite{ames2014rapidly, nguyen2015optimal, galloway2015torque}. Subsequent work incorporated inverse dynamics (ID) directly into the optimization variables \cite{reher2019inverse}. This strategy optimizes joint accelerations subject to both the full body dynamics and the CLF inequality. The ID-CLF-QP formulation was shown to improve numerical conditioning and allows for faster convergence \cite{reher2019inverse}. 

Despite these advances, the application of CLF-QP methodologies to underactuated redundant soft robots remains unexplored. In this paper, we present a general framework for constructing CLF-QP controllers tailored to control of underactuated soft and soft-rigid robots. We summarize the following contributions: 
\begin{itemize}
    \item A CLF-QP control framework adapted for soft robots.
    \item A Soft ID-CLF-QP controller that accounts for underactuation of soft robots.
    \item Validation on new benchmark soft-rigid robot systems.
    \item Comparison with other control approaches.
    \item Open source, flexible Python implementations for testing on new systems and with new controllers \footnote{\url{https://github.com/CyPhiLab/tendon_clf}}.
\end{itemize}

\section{Soft Robot Model Systems}\label{sec:sim}

\begin{figure}[t]
    \centering
    \includegraphics[width=0.47\textwidth]{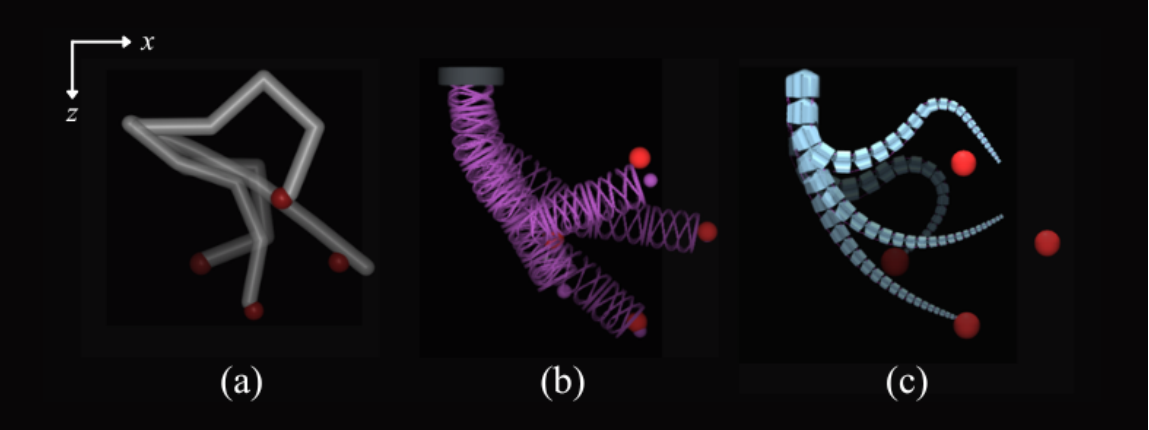}
    \caption{Results of our proposed Soft ID-CLF-QP control approach operating on various soft robot architectures to solve the task-space control problem. The robot platforms are: (a) Finger (L = 0.24 m), (b) Helix (L = 0.45 m), and (c) SpiRob (L = 0.50 m).}
    \label{experiment}
\end{figure}

\subsection{Robot Dynamics}
For all of our model systems, we represent the dynamics in the following standard form:
\begin{equation}
M(q)\ddot{q} + C(q,\dot{q})\dot{q} + D(q)\dot{q} + K(q) + g(q) = Bu
\label{dynamics}
\end{equation}
where  
\begin{itemize}
    \item $q \in \mathbb{R}^{n}$ is the vector of joint positions, with $\dot{q}, \ddot{q} \in \mathbb{R}^{n}$ denoting velocities and accelerations,
    \item $M(q) \in \mathbb{R}^{n \times n}$ is the positive-definite inertia matrix,  
    \item $C(q,\dot{q}) \in \mathbb{R}^{n \times n}$ represents Coriolis and centrifugal effects,  
    \item $D(q) \in \mathbb{R}^{n \times n}$ is the joint-space damping matrix and $K(q) \in \mathbb{R}^{n \times n}$ is the force from the stiffness potential.
    \item $G(q) \in \mathbb{R}^{n}$ is the gravity force vector,
    \item $B \in \mathbb{R}^{n \times m}$ is the actuation matrix that maps control inputs into joint torques,  
    \item $u \in \mathbb{R}^{m}$ is the vector of control inputs.  
\end{itemize}
Note that our controllers do not explicitly account for collisions between adjacent links if they occur. 

\subsection{Robots}
We compare three robotic platforms with varying levels of underactuation.
\subsubsection{Tendon-Driven Finger}
The tendon-driven finger (Fig. \ref{experiment}a) is modeled as four rigid links connected in series and actuated by two tendon pairs, each tendon pair spans two adjacent joints.
\subsubsection{Trimmed Helicoid Soft Continuum}
The Helix robot (Fig. \ref{experiment}b) used in this work follows the soft–rigid hybrid robot presented in \cite{patterson2025design}, which was based on a similar continuum robot \cite{guan2023trimmed}. The robot is cable-driven and actuated by nine motors mounted at the base. Helix's body consists of three modules. To represent soft robotic deformation, we forgo kinematic assumptions such as piecewise constant curvature \cite{webster2010design} or rod based techniques such as Cosserat rod theory \cite{trivedi2008geometrically}. Instead, we adopt a discretized rigid multibody approximation of the soft structure, similar to the approach utilized in \cite{zwane2024learning}. The continuum body is approximated by a finite number of rigid segments connected by compliant joints, resulting in a structured, finite-dimensional model amenable to simulation in standard robotics packages. The helix robot has 3 modules, which are each discretized into 4 segments. Each joint between adjacent segments has 3 DOFs, resulting in a total of 36 DOFs.

\subsubsection{Logarithmic Spiral Robot}
SpiRob (Fig. \ref{experiment}c) is a logarithmic spiral-shaped continuum robot inspired by biological appendages such as octopus arms and elephant trunks \cite{wang2025spirobs}. The robot is cable-driven, with three cables routed along the deformable backbone. SpiRob is highly underactuated, with only three actuators controlling a 27-DOF continuum structure.

\section{Control Methodology}\label{sec:methods}
We first perform input–output linearization to derive linear task-space dynamics. CLF-QP controllers are then developed that ensures Lyapunov stability and explicitly incorporate actuator constraints within a convex optimization framework.

\subsection{Input--Output Linearization}

Let $x = [q, \dot{q}]^\top$. Equation~\eqref{dynamics} can be written in the control-affine form:
\begin{equation}
\dot{x} = f(x) + g(x)u,
\end{equation}

Let $y$, $y_{\text{ref}}(t)$, and $e = y - y_{\text{ref}}(t)$ be the end-effector position, the target trajectory, and the end-effector position error in Cartesian space  ($y, y_{\text{ref}} \in \mathbb{R}^3$), respectively. The manipulator Jacobian $J(q) \in \mathbb{R}^{3 \times n}$ maps the joint velocities to the end-effector velocity: 
\begin{equation}
\dot{y} = J(q)\dot{q},
\label{2}
\end{equation}
Then taking the second time derivative of $e$ yields:
\begin{equation}
\ddot{e} = \ddot{y} - \ddot{y}_{\text{ref}} = L_f^2 y + L_g L_f y  u- \ddot{y}_{\text{ref}}(t) = \mu,
\end{equation}
where $L_f$ and $L_g$ denote Lie derivatives along $f(x)$ and $g(x)$. We then select a control input which renders the linearized task-space error dynamics:
\begin{equation}
u = \big(L_g L_f y\big)^{-1}
\left(-L_f^2 y + \mu + \ddot{y}_{\text{ref}}\right)
\quad \Longrightarrow \quad
\ddot{e} = \mu,
\label{io_linear}
\end{equation}

Let $\eta = [ e, \dot{e}]^\top$, the linearized task error dynamics are achieved:
\begin{equation}
\dot{\eta} = F_{\mathrm{L}}\eta + G_{\mathrm{L}}\mu,
\label{task_space_dynamics}
\end{equation}
where
\begin{equation}
F_{\mathrm{L}} = \begin{bmatrix} 0 & I \\[2pt] 0 & 0 \end{bmatrix}, 
\qquad
G_{\mathrm{L}} = \begin{bmatrix} 0 \\[2pt] I \end{bmatrix}.
\end{equation}

\subsection{CLF--QP  Controller}

To drive the error state $\eta$ to zero, we define the function $V(\eta)$ as a rapidly exponentially stabilizing control Lyapunov function (RES-CLF) \cite{ames2014rapidly} for the linearized task-space dynamics in~\eqref{task_space_dynamics}:

\begin{equation}
V = \eta^{\top}
\begin{bmatrix}
\frac{1}{\varepsilon} I & 0 \\
0 & I
\end{bmatrix}
P
\begin{bmatrix}
\frac{1}{\varepsilon} I & 0 \\
0 & I
\end{bmatrix}
\eta = \eta^{\top} P_{\varepsilon} \eta,
\end{equation}

\noindent where $P \succ 0$ is the solution of the continuous-time algebraic Riccati equation:
\begin{equation}
F_{\mathrm{L}}^\top P + P F_{\mathrm{L}} - P G_{\mathrm{L}} R^{-1} G_{\mathrm{L}}^\top P + Q = 0,
\label{eq:care}
\end{equation}
with design weights $Q = Q^\top \succeq 0$ and $R = R^\top \succ 0$.

The derivative of $V$ with respect to time is:
\begin{equation}
\dot{V}(\eta, \mu) = \eta^\top \big( F_{\mathrm{L}}^\top P_{\varepsilon} + P_{\varepsilon} F_{\mathrm{L}} \big)\eta + 2\eta^\top P_{\varepsilon} G_{\mathrm{L}} \mu,
\label{3}
\end{equation}

To enforce rapid exponential convergence, we formulate a QP that takes the RES-CLF condition: 
\[
\dot V(\eta,\mu) \le -\frac{1}{\varepsilon} V(\eta),
\]
as an inequality constraint.
A relaxation variable $\delta \ge 0$ is introduced to guarantee feasibility of the QP in the presence of actuator saturation, dynamic coupling, or model mismatch. The CLF inequality is therefore enforced in the softened form:
\[
\dot V(\eta,\mu) \le -\frac{1}{\varepsilon} V(\eta) + \delta,
\]
with $\delta$ heavily penalized in the objective to encourage strict satisfaction of the Lyapunov decrease condition whenever feasible.

To regulate the task space, we penalize deviations of the task error acceleration $\mu =J\ddot{q} + \dot{J}\dot{q}-\ddot{y}_{\text{ref}}$ from a PD reference of the form  $ \mu_{\text{ref}} = - K_d\dot{e}- K_pe$, where $K_p$ and $K_d$ are positive definite gain parameters. We choose the derivative gain as $K_d =2 \sqrt{K_p}$ to encourage critically damped behavior. Actuators are constrained to obey the input output linearizing controller ~\eqref{io_linear} and input bounds are incorporated as additional inequality constraints. This is a standard CLF-QP controller. We drop dependencies for brevity.

\noindent\rule{\linewidth}{0.4pt}
CLF-QP:
\begin{equation}
    \begin{aligned}
        \argmin_{u, \mu, \delta} \quad & w_1 \big\|{\mu} - \mu_{\text{ref}} \big\|^2 + \rho \delta^2 \\
        \textrm{s.t.} \quad & \dot{V} \leq - \frac{1}{\epsilon} V + \delta, \\
        & u = (L_g L_f y)^{-1} (-L_f^2 y + \mu + \ddot{y}_{\text{ref}}), \\
        & u_{\min} \leq u \leq u_{\max}.
    \end{aligned}
\end{equation}
\noindent\rule{\linewidth}{0.4pt}

\subsection{Soft ID--CLF--QP  Controller}

While the CLF--QP controller works in many scenarios, we found that it failed for some systems that were highly redundant with respect to the task space because the controller is able to excite zero dynamics to produce instability in the nullspace of the task. Instead of input-output linearizing, the ID--CLF--QP controller is instead able to constrain control inputs based on the whole-body dynamics, while simultaneously incentivizing the task space to converge and regularizing accelerations in the state space, preventing such degenerate behaviors. 

In the ID--CLF--QP, the full system dynamics are imposed as an affine equality constraint according to~\eqref{dynamics} so that the optimal task-space acceleration $\mu$ is realized through inverse dynamics \cite{reher2019inverse}. To damp redundant joint motions near equilibrium, a null-space acceleration term $\ddot{q}_{\mathrm{null}} = N\ddot{q}$ is included and kept close to a reference 
$\ddot{q}_{\mathrm{null}}^{\mathrm{ref}} = - D\, N\, \dot{q}$, where $N = I - J^+ J$ denotes the null-space projection matrix, $D$ is the null-space damping parameter, and $J^+$ is the Moore-Penrose pseudoinverse. Joint accelerations $\ddot{q}$ are also penalized to prevent excessively large or ill-posed solutions resulting from the task-space mapping.  A critical finding in this work is that simply encoding our dynamics constraints as written in the form of~\eqref{dynamics} leads to poor performance and stability issues. This is because it constrains the control input to both achieve the task (via the CLF constraint) and to generate the full set of accelerations to fulfill the inverse dynamics, which is often not achievable in systems with a high degree of underactuation. Instead, we propose to implement the hard inverse dynamics constraint on the actuated subspace, while allowing a soft constraint on the unactuated subspace. From \cite{pustina2024input}, for thread-like actuators (such as tendons and most other actuators in soft robotics), there is always a change of coordinates $T(q) = [T_\mathrm{a}(q), T_\mathrm{u}(q)]^{\top}$ such that the state can be partitioned into actuated and unactuated coordinates: $T(q)q = [q_\mathrm{a}, q_\mathrm{u}]^{\top}.$ We then apply our change of coordinates to the dynamics to get a \textit{collocated form} \cite{pustina2024input} where the actuated dynamics can be written as 
\begin{equation}
    \bar{M}_\mathrm{a}\ddot q_\mathrm{a} + \bar{h}_\mathrm{a}(q,\dot q) = u,
    \label{actuated}
\end{equation}
with $\bar{M}_\mathrm{a} = T_\mathrm{a}^{-\top}MT_\mathrm{a}^{-\top}$, $\bar{h}_\mathrm{a}(q,\dot q) = T_\mathrm{a}^{-\top}h(q,\dot q)T_\mathrm{a}^{-\top}$, and $h(q,\dot q) = C(q,\dot{q})\dot{q} + D(q)\dot{q} + K(q) + g(q)$.
Without loss of generality, if the input matrix is constant, $T$ is a constant and $T_\mathrm{a} = B^{\top}$. This fact can be used to efficiently to efficiently calculate T online.

This controller causes the accelerations on the actuated coordinates to verify the stability conditions imposed by the CLF while also generating physically viable behavior through the actuated dynamics constraint. Thus, it inherits the stability properties of past presentations of the CLF-QP \cite{galloway2015torque} and ID-CLF-QP \cite{reher2019inverse} with rapid exponential stabilization and actuator saturation. Critically, the stability of the closed loop system is reliant on stable zero dynamics. In the soft robotics case, a sufficient condition for stable zero-dynamics is that the stiffness in the unactuated coordinates dominates gravity (or that $\frac{\partial K(q)}{\partial q_\mathrm{u}} > - \frac{\partial G(q)}{\partial q_\mathrm{u}}$) \cite{pustina2022feedback}, a condition fulfilled by many soft robotic systems.

\noindent\rule{\linewidth}{0.4pt}
Soft ID-CLF-QP:
\begin{equation}
    \begin{aligned}
        \argmin_{\ddot{q}, u, \delta} \quad & w_1 \big\|{\mu} - \mu_{\text{ref}} \big\|^2 + w_2 \|\ddot{q}\|^2 + w_3 \|u\|^2 \\ &+ w_4\|\ddot{q}_\text{null} - \ddot{q}_{\mathrm{null}}^{\mathrm{ref}}\|  + \rho \delta^2  \\
        \textrm{s.t.} \quad & \dot{V} \leq - \frac{1}{\epsilon} V +  \delta, \\
        &  \bar{M}_\mathrm{a}\ddot q_\mathrm{a} + \bar{h}_\mathrm{a}(q,\dot q) = u,\\
        & u_{\min} \leq u \leq u_{\max}.
    \end{aligned}
\end{equation}
\noindent\rule{\linewidth}{0.4pt}


\section{Validation}\label{sec:results}

To quantify the performance of the ID-CLF-QP controller, we developed the robot models and simulated them in MuJoCo \cite{todorov2012mujoco}. The evaluation focuses on the robot's ability to reach various set positions and to track a trajectory in space. The coordinate axes used to reference the target positions are shown in Fig.\ref{experiment}. The controllers are then compared against a set of baseline controllers described in section A below. All controllers are implemented in Python and optimization problems are solved using the cvxpy library \cite{cvxpy}. The control gains used in the experiments are included in Table \ref{tab:controller_params}.

\subsection{Baseline}
\subsubsection{Impedance Control (IC)}
IC is a classic robot control strategy that regulates the dynamic relationship between the end-effector motion and external interaction forces \cite{khatib1987a}. IC enables trajectory tracking by prescribing compliant second-order error dynamics in a fully actuated task-space setting through a task-space wrench $f$ given by:
\begin{equation}
    f = \Lambda(q)\ddot{y}+h(q,\dot{q}) ,
\label{eq:impedance_acc}
\end{equation}
with
\begin{equation}
    (\ddot{y} - \ddot{y}_{\text{ref}}) + K_d (\dot{y} - \dot{y}_{\text{ref}}) + K_p (y - y_{\text{ref}}) = 0,
\label{impedance_acc}
\end{equation}

\noindent 
where
$\Lambda(q)=\big(J(q)M^{-1}(q)J^\top(q)\big)^{-1}$ and 
$h(q,\dot{q})=J^{+\top}C(q,\dot{q})\dot{q}-\Lambda(q)\dot{J}(q,\dot{q})\dot{q}$
are the task-space inertia  and  Coriolis matrices, respectively. We directly cancel stiffness, damping, and gravitational forces. Therefore, the controller is as follows. We drop dependencies for brevity.

\noindent\rule{\linewidth}{0.4pt}
IC:
\begin{equation}
    u = \text{clamp}\Big(B^{+}(J^{\top}f + D\dot{q} + Kq + G), u_{\text{min}}, u_{\text{max}}\Big).
\label{IC}
\end{equation}
\noindent\rule{\linewidth}{0.4pt}

In this vanilla IC formulation, since joint space dynamics are underactuated, \eqref{IC} finds actuator commands that best approximate the joint torques required to generate the desired task-space force in a least-squares sense. 

\subsubsection{Underactuated Impedance Control (UIC)}
To generate torque commands more consistent with underactuated dynamics, \cite{mistry2012operational} proposed a method that removes input torque components in unactuated directions via null-space projection. We propose a similar method as a benchmark here:
\begin{equation}
    Bu = J^{\top}(q)f + N\tau_{\text{null}}
\label{uic}
\end{equation}

\noindent The underactuation imposes a constraint on the admissible joint torques:
\begin{equation}
    J^{\top}(q)f + N\tau_{\text{null}} =
    B^{+}B\Big(J^{\top}(q)f + N\tau_{\text{null}}\Big),
\label{null}
\end{equation}

\noindent
where $I_p = B^{+}B$ denotes the least-squares projection onto the actuated joint subspace. 
From \eqref{uic} and \eqref{null}, the null-space torque $\tau_{\text{null}}$ is derived as:
\begin{equation}
    \tau_{\text{null}} =
    -\Big[(I - I_p)N\Big]^{+}
    (I - I_p)J^{\top}(q)f.
\label{null_tau}
\end{equation}

\noindent The UIC law is then formulated as:

\noindent\rule{\linewidth}{0.4pt}
UIC:
\begin{equation}
\begin{aligned}
    \tau &= \Big(I - N[(I - I_p)N]^+\Big) J^{\top}f + D\dot{q} + Kq + G, \\
    u &= \text{clamp}\Big(B^{+}\tau, u_{\text{min}}, u_{\text{max}}\Big).
\end{aligned}
\label{UIC}
\end{equation}
\noindent\rule{\linewidth}{0.4pt}

\subsubsection{Augmented-QP Impedance Control}
Since vanilla task-space control techniques do not explicitly account for actuator bounds, the control objective can be wrapped in a QP to impose input bounds. We leverage the same underactuated inverse dynamics constraint as the one implemented for Soft ID-CLF-QP, and we define for
\begin{equation}
     \mu =J\ddot{q} + \dot{J}\dot{q}-\ddot{y}_{\text{ref}}
\end{equation}

\noindent\rule{\linewidth}{0.4pt}
IC-QP:
\begin{equation}
    \begin{aligned}
        \argmin_{\ddot{q}, u} \quad & w_1 \big\|{\mu} - \mu_{\text{ref}} \big\|^2 + w_2 \|\ddot{q}\|^2 + w_3 \|u\|^2 \\ &+ w_4\|\ddot{q}_\text{null} - \ddot{q}_{\mathrm{null}}^{\mathrm{ref}}\|  \\
        \textrm{s.t.} \quad
        &  \bar{M}_\mathrm{a}\ddot q_\mathrm{a} + \bar{h}_\mathrm{a}(q,\dot q) = u,\\
        & u_{\min} \leq u \leq u_{\max}.
    \end{aligned}
\end{equation}
\noindent\rule{\linewidth}{0.4pt}
This structure is very similar to the Soft ID-CLF-QP problem. The only missing feature is indeed the CLF term.

\begin{table}[t]
\centering
\caption{Controller parameter settings for each robot platform.}
\label{tab:controller_params}
\resizebox{0.95\linewidth}{!}{%
\begin{tabular}{|c|c|c|c|c|c|c|c|c|}
\hline
\textbf{Robot} & \textbf{Controller} & $K_p$ & $\epsilon$ & $w_1$ & $w_2$ & $w_3$ & $w_4$ & $\rho$ \\
\hline
\multirow{5}{*}{Finger} 
& CLF-QP     & 500.0  & 0.05 & 1 & --  & -- & -- & 1000 \\
& ID-CLF-QP  & 500.0  & 0.03 & 1 & 0.2  & 0.2 & 0.1 & 1000 \\
& IC         & 500.0  & 0.03 & -- & -- & -- & -- & -- \\
& UIC        & 500.0  & 0.03 & -- & -- & -- & -- & -- \\
& IC-QP      & 1500.0 & 0.03 & 1 & 0.02 & 0.2 & -- & -- \\
\hline
\multirow{5}{*}{Helix} 
& CLF-QP     & 500.0  & 0.05 & 1 & -- & -- & -- & 1000 \\
& ID-CLF-QP  & 500.0  & 0.05 & 1 & 0.1 & 0.2 & 0.1 & 1000 \\
& IC         & 2000.0 & 0.05 & -- & -- & -- & -- & -- \\
& UIC        & 1000.0 & 0.05 & -- & -- & -- & -- & -- \\
& IC-QP      & 500.0  & 0.05 & 1 & 0.1 & 0.2 & -- & -- \\
\hline
\multirow{5}{*}{SpiRob} 
& CLF-QP     & 500.0  & 0.01 & 1 & -- & -- & -- & 1000 \\
& ID-CLF-QP  & 500.0  & 0.01 & 1 & 0.2 & 0.5 & 0.1 & 1000 \\
& IC         & 500.0  & 0.01 & -- & -- & -- & -- & -- \\
& UIC        & 500.0  & 0.01 & -- & -- & -- & -- & -- \\
& IC-QP      & 2000.0 & 0.01 & 1 & 0.2 & 0.5 & -- & -- \\
\hline
\end{tabular}}
\end{table}

\subsection{Experiments}
\subsubsection{Set Point Regulation}
In this experiment, we evaluate all three robots' ability to reach and stabilize set points in space ($\dot{e} = \ddot{e} = 0$). The experimental setup is illustrated in Fig.~\ref{experiment}, where the end-effector tracks a target point (red dot). The set points are distributed along an elliptic path in the $xz$-plane, parameterized as:
\begin{equation}
\label{ellipse}
\begin{aligned}
x &= a\cos\theta \cos\phi 
     - \big(b\sin\theta - c\big)\sin\phi, \\
z &= a\cos\theta \sin\phi 
     + \big(b\sin\theta - c\big)\cos\phi,
\end{aligned}
\end{equation}
where $L$ is the total robot length, $a = \frac{L}{3}$ and $b = \frac{L}{8}$ define the semi-major and semi-minor axes of the ellipse, respectively, and $\phi = \frac{\pi}{4}$ is the clockwise rotation angle about the $y$-axis. The angle $\theta \in \{0, 0.5, 1,1.5\}\pi$ specifies the four target points, and the robot base is located at $(0,0,0)^\top$. The parameter $c$ is the distance from the robot base to the ellipse center, defined as $c = L - b$ for the \textit{Finger} and \textit{Helix} robots, and $c = \frac{3L}{4} - b$ for the \textit{SpiRob}. All simulations are set to stop after 10 seconds.

\subsubsection{Trajectory Tracking}

In this experiment, each robot tracks a continuous reference trajectory defined along the same elliptic path introduced in the set point experiment (see \eqref{ellipse}). 
The trajectory is parameterized by:
\begin{equation}
\theta(t) = \omega t,
\end{equation}
\noindent where $\omega \in \{0.1, 0.2, 0.3, 0.4, 0.5\}\pi$  (rad/s) is the angular frequency. Tracking experiments are terminated after completing two full cycles ($4\pi/\omega$~seconds).

\subsection{Results}

\begin{figure}
    \centering
    \includegraphics[width=0.95\linewidth]{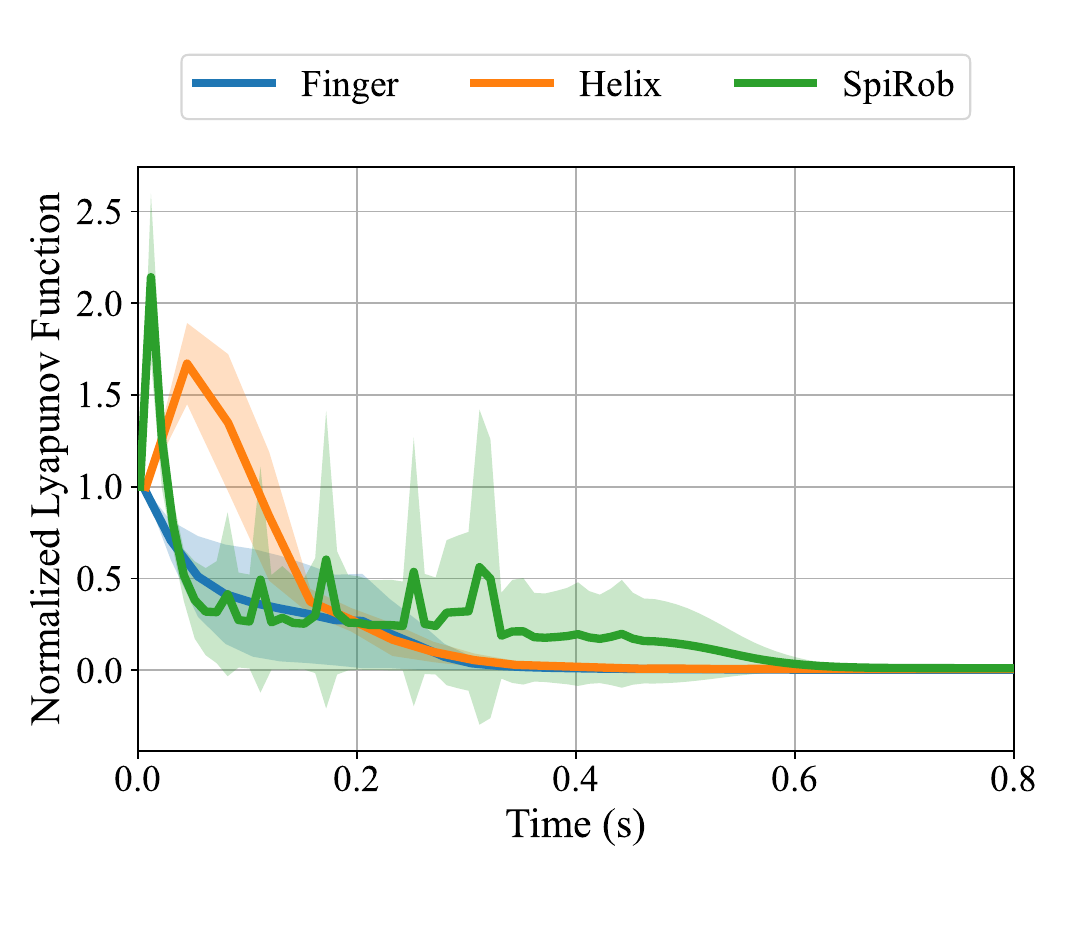}
    \caption{Lyapunov functions over time for set points tracking experiment.}
    \label{clf_set}
\end{figure}

\begin{figure}
    \centering
    \includegraphics[width=0.95\linewidth]{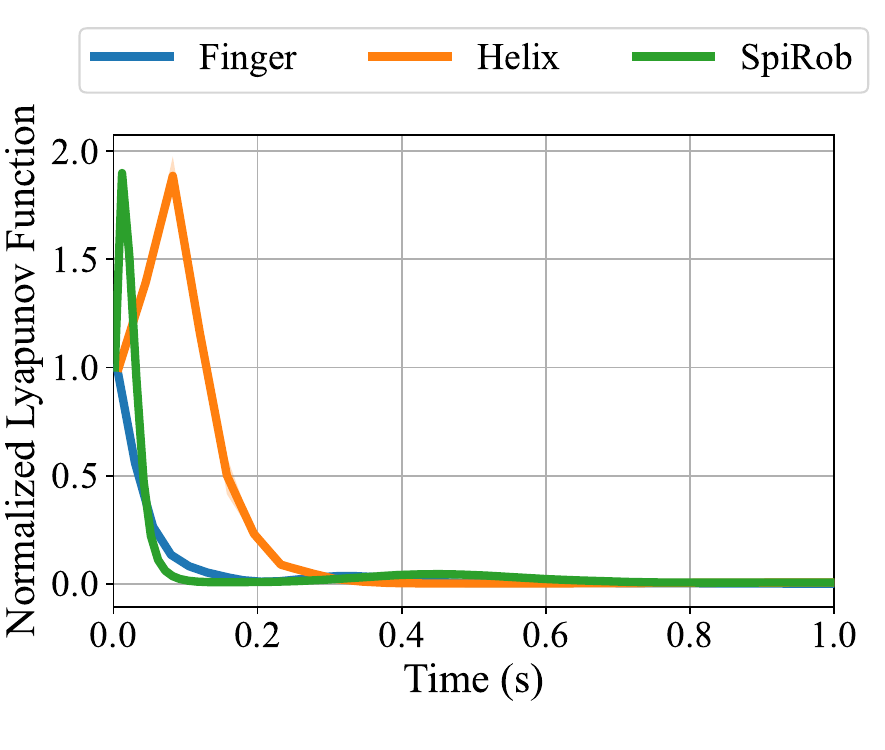}
    \caption{Lyapunov functions over time for trajectory tracking experiment.}
    \label{clf_tracking}
\end{figure}

\begin{figure}
    \centering
    \includegraphics[width=0.95\linewidth]{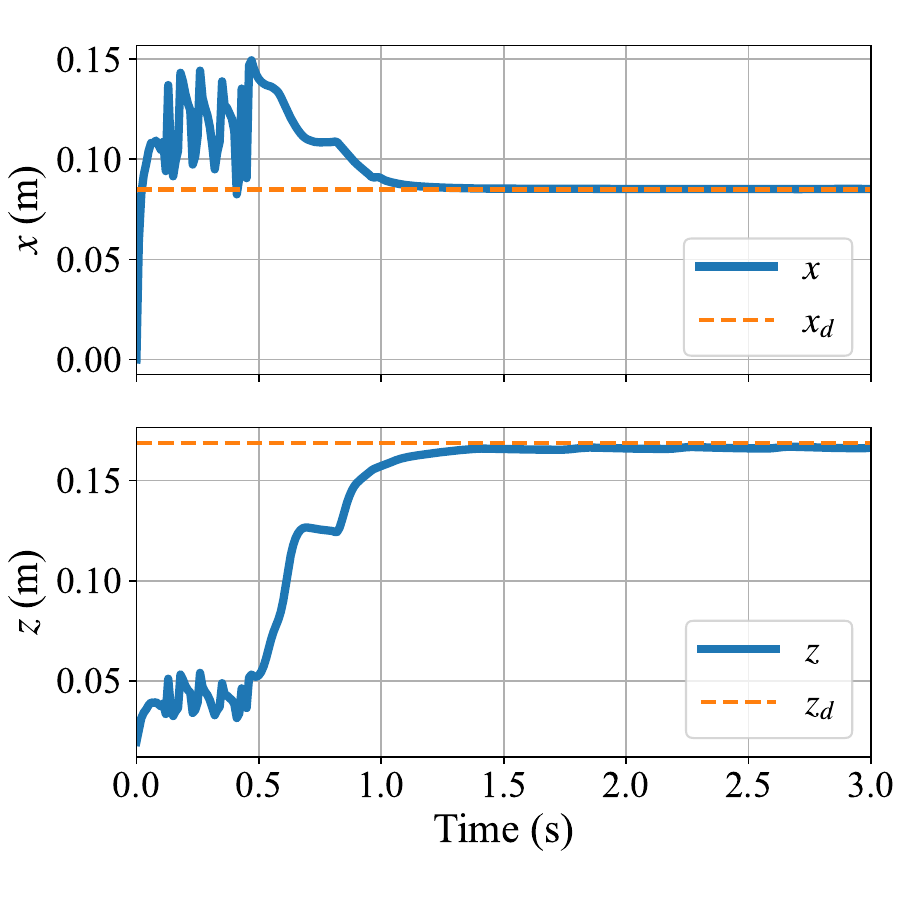}
    \caption{Task-space convergence of the SpiRob during a set-point regulation experiment in the $x$--$z$ plane. The variables $x_d$ and $z_d$ denote the target point coordinates.}
    \label{spirob_error}
\end{figure}

The effectiveness of our proposed controller is demonstrated through the convergence of the Lyapunov functions in both set point reaching and trajectory tracking experiments (Figs.~\ref{clf_set} and \ref{clf_tracking}). The Lyapunov functions are normalized by their initial value $V(\eta(0))$ to allow fair comparison across different robot platforms. We also plot task space convergence during a characteristic experiment for a Spirob set point trial (Fig. \ref{spirob_error}).

A quantitative benchmark comparison is presented in Table~\ref{tab:combined_benchmark}. In the set point regulation experiments, we report the mean final error and standard deviations. In the trajectory tracking experiments, we report the mean squared error (MSE) and and standard deviations. Because these are deterministic controllers, these distributions correspond to performance for different set points and trajectories rather than to stochasticity. 

The proposed Soft ID-CLF-QP controller achieves the best performance on four out of six benchmarks, and it is second in the other benchmarks, indicating reliable performance across this range of settings. In general, the other approaches typically struggle in one or more settings. For example, CLF-QP often manages reasonable performance but fails to run for the Helix robot. Similarly, both closed form controllers (IC and UIC) fail to converge on the logarithmic spiral robot, while IC-QP is able to somewhat move the robot in the correct direction, but ultimately fails to converge.

\begin{figure}
    \centering
    \includegraphics[width=0.9\linewidth]{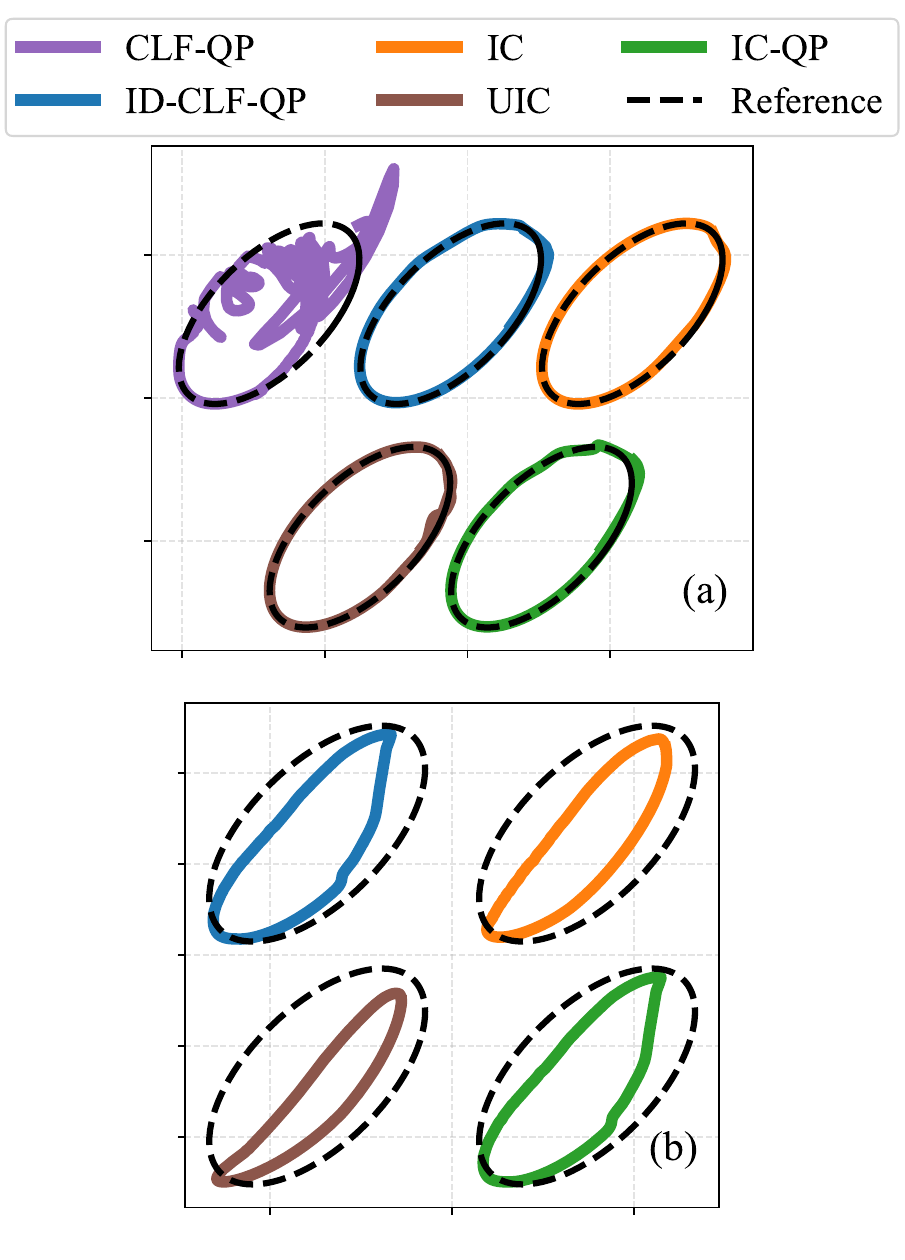}
    \caption{Task-space trajectory tracking performance during the second cycle across controllers for Finger (a) and Helix (b) for $\omega = 0.2\pi$ rad/s.}
    \label{tracking}
\end{figure}

Figs.~4(a) and~4(b) illustrate the tracking performance of the robots under the tested control schemes during the second cycle of motion. As is observed, all controllers experience degradation in tracking accuracy when the end–effector moves into difficult regions. As a result, the generated motion deviates from the reference trajectory, particularly along directions that require large joint accelerations. 

\begin{table}[t]
\centering
\caption{Combined benchmark comparison for set point (SP) and trajectory tracking (TT) across three robot platforms.}
\label{tab:combined_benchmark}
\resizebox{0.99\linewidth}{!}{%
\begin{tabular}{|c|c|c|c|}
\hline
\textbf{Robot} & \textbf{Controller} & \textbf{Final Error (SP) (cm)} & \textbf{TT--MSE ($\mathbf{cm^2}$)} \\
\hline
\multirow{6}{*}{Finger} & CLF-QP & 0.75 ± 1.05 & 6.90 ± 0.80 \\
& ID-CLF-QP (ours) & 0.08 ± 0.09 & \textbf{0.53 ± 0.26} \\
& IC & 0.08 ± 0.14 & 1.26 ± 0.73 \\
& UIC & 0.22 ± 0.39 & 1.00 ± 0.51 \\
& IC-QP & \textbf{0.07 ± 0.12} & 1.26 ± 0.71 \\
\hline
\multirow{6}{*}{Helix} & CLF-QP & Failed Convergence & Failed Convergence \\
& ID-CLF-QP (ours) & \textbf{2.48 ± 1.20} & 24.14 ± 9.04 \\
& IC & 2.66 ± 1.40 & \textbf{21.89 ± 6.08 }\\
& UIC & 3.39 ± 2.12 & 31.70 ± 7.48 \\
& IC-QP & \textbf{2.48 ± 1.20} & 24.14 ± 9.04 \\
\hline
\multirow{6}{*}{SpiRob} & CLF-QP & 4.23 ± 3.98 & 255.33 ± 186.38 \\
& ID-CLF-QP (ours) & \textbf{3.50 ± 3.71} & \textbf{19.51 ± 5.66} \\
& IC & Failed Convergence & Failed Convergence \\
& UIC & Failed Convergence & Failed Convergence \\
& IC-QP & 8.93 ± 7.80 & 157.64 ± 6.48\\
\hline
\end{tabular}}
\end{table}
\section{Discussion}

Our proposed control approaches achieve effective performance among the competing control schemes, with various tradeoffs between tracking accuracy and computational cost.

The CLF-QP framework takes advantage of linearized task-space error dynamics to construct a simple but efficient candidate Lyapunov function based on the LQR value (cost-to-go) function. The base CLF-QP controller generalizes well to certain types of robots, where the unactuated dynamics are inherently subject to enough state constraints such that the zero dynamics are not excited, but without explicitly regulating the acceleration $\ddot q$, this controller does not converge for situations where the joints of the robot are not constrained - such as an soft robot with a compressible/extensible backbone or one modeled with discretized rods connected by universal joints such as the Helix robot in this work. There is nothing in such a scenario preventing the optimizer from choosing arbitrary accelerations, whether or not they are compatible with the robot's task-redundant dynamics. 

We addressed this problem by adapting the ID-CLF-QP controller from \cite{reher2019inverse}, which explicitly regulates $\ddot q$, preventing such instability. However, the optimization does not converge for the direct application of the ID-CLF-QP controller to highly underactuated soft robots, such as those modeled here, because the controller is only capable of directly regulating accelerations on the so-called collocated coordinates, while it cannot directly regulate unactuated coordinates. Therefore, drawing from \cite{pustina2024input} we proposed a change of coordinates and constrained only the actuated set of our transformed coordinate system to obey our inverse dynamics constraint. This effectively allows the optimizer to choose the best \textit{controllable} accelerations that realize the task-space goal. 

Uniquely, the Soft ID-CLF-QP preserves stability and achieves good performance across tasks. Critically, it is the only controller that reliably solves the experiments for all three environments, and it the only one to preserve even mediocre performance for tracking for the logarithmic spiral robot. We believe this flexibility inherits from the unique structure of the controller, which leverages the physical model imbued by inverse dynamics, CLF-based stability properties, and robustness due to operating directly on the actuated coordinates. This allows for a controller that does not fight against the natural dynamics and underactuated structure, but rather attempts to leverage these properties. Each component of this is critical, as the other benchmarks can be seen as testing the various individual components of the controller. 


One interesting limitation of the controllers presented in this work is that they all utterly fail to solve the tracking problem for the logarithmic spiral robot. We believe this is due to the interaction of a number of factors including the complex, path dependent kinematics and the high degree of underactuation (27 joints to 3 actuators). Solving the control problem more effectively likely requires either a more sophisticated control approach that uses longer horizon planning, such as reinforcement learning or nonlinear MPC, or a control approach that simplifies the planning burden by resorting to pre-specified motion primitives. 

One caveat related to the algorithms presented in this work is that all are implemented in Python. Therefore, reported computation times should be considered conservative. It is well known in the literature that similar optimization-based controllers can run at roughly 1kHz rates when implemented in C++ on real robots \cite{bledt2018cheetah, reher2019inverse}. 

\section{Conclusion}
\label{sec:conclusion_future_work}

Through this research, we have developed a highly general optimization-based framework for using control Lyapunov functions with strong convergence properties for task based control of underactuated soft robots. Within this framework, we introduced a novel ``Soft'' ID-CLF-QP framework, which uses a change of coordinates to produce a relaxation of the inverse dynamics constraints along the unactuated coordinates. Combined with the stability properties of the standard CLF-QP framework, this results in a controller that leverages the physical and dynamical structure of the problem to achieve reliable performance across a variety of environments. We also provided a rigorous comparison to a variety of other control approaches across multiple benchmark environments.
Future work includes augmenting the controller with control barrier functions (CBFs) to enhance safety during interaction with humans and the environment \cite{nguyen2016exponential, dickson2025safe, patterson2024safe}. Another potential direction is to investigate task-centric control for soft robotic grasping \cite{huang2025grasping}. 

To further validate the simulation results, the controller will be deployed on a physical hardware platform. A key challenge will be real-time state estimation. To overcome this difficulty, we plan to investigate vision-based sensing approaches—such as pixel-to-torque mappings inspired by \cite{lee2024pixelstorqueslinearfeedback}—combined with motor encoder measurements to perform reliable state estimation on hardware.


\bibliographystyle{ieeetran}
\bibliography{bib}

\end{document}